\documentclass{article}

\usepackage{microtype}
\usepackage{graphicx}
\usepackage{float}
\usepackage{subcaption}
\usepackage{booktabs} % for professional tables
\usepackage{tabularx}
\usepackage{array}
\usepackage{fvextra}
\usepackage{alltt}
\usepackage{adjustbox}
\usepackage{caption}
\usepackage{hyperref}

\usepackage[preprint]{icml2026}

\usepackage[scaled=0.95]{inconsolata}

\usepackage{amsmath}
\usepackage{amssymb}
\usepackage{mathtools}
\usepackage{amsthm}

\usepackage[capitalize,noabbrev]{cleveref}

\theoremstyle{plain}

\theoremstyle{definition}

\theoremstyle{remark}

\usepackage[textsize=tiny]{todonotes}

\icmltitlerunning{The DeepSpeak-Agentic Dataset}

\begin{document}

\twocolumn[
  \icmltitle{The \emph{DeepSpeak-Agentic} Dataset}



 \icmlsetsymbol{equal}{*}
\begin{icmlauthorlist}
  \icmlauthor{Sarah Barrington}{equal,ucb}
  \icmlauthor{Maty Bohacek}{equal,stanford}
  \icmlauthor{Hany Farid}{equal,ucb}
\end{icmlauthorlist}
\icmlaffiliation{ucb}{University of California, Berkeley, USA}
\icmlaffiliation{stanford}{Stanford University, USA}
\icmlcorrespondingauthor{Sarah Barrington}{sarah.barrington@berkeley.edu}
\icmlcorrespondingauthor{Maty Bohacek}{maty@stanford.edu}
\icmlcorrespondingauthor{Hany Farid}{hfarid@berkeley.edu}
\icmlkeywords{Agents, Avatars, Human-AI Interaction, Dataset}
\vskip 0.3in
]



\printAffiliationsAndNotice{\icmlEqualContribution}  % no special notice (required even if empty)

\begin{abstract}
  We present {\em DeepSpeak-Agentic}, a dataset of videos comprising over $37$ hours of semi-structured conversations between a human and an embodied AI agent. We use this dataset to evaluate the automatic forensic identification (audio, video, or text) of AI agents, study the nature of human-agent interactions, and provide a benchmark for future advances in the large-language models and AI-generated voices and faces that power embodied AI agents. We also contribute a scalable data-capture system that creates agents, automatically pairs them with human crowd workers, records audiovisual conversations across specified scenarios, and identifies and separates the human and agent in the combined stream. 
\end{abstract}

%%%%%%%%%%%%%%%%%%%%%%%%%%%%%%%%%%%%%%%%%%%%%%%%%%%%%%%%%%%%%%%%%%%%%%%%%%%%%%%

\section{Introduction}

In 2024, Zoom's CEO Eric Yuan envisioned a future where an agentic AI clone would join meetings on our behalf~\cite{verge24}. Just $16$ months later, Yuan delivered Zoom's 2025 Q1 earnings with his AI clone. This clone, however, was prerecorded and was only able to deliver the prepared opening remarks, and not respond in real time to questions. In response to an analyst's comment ``Nice avatar.'', Yuan responded. ``Thank you. Appreciate it. Next earnings call will be much better''~\cite{zoom25} (presumably Yuan was referring to the AI clone, not Zoom's earnings).

From a technical perspective, Yuan's prediction was correct. This past year has seen tremendous progress in real-time, embodied agents powered by large language models (LLMs), synthetic voices, and visual avatars, getting us closer to Yuan's vision. Indeed, today, there are multiple commercial offerings that allow humans to converse with agents (e.g.,~\url{https://anam.ai}, \url{https://heygen.com}, and \url{https://tavus.com}). If this trajectory continues, Yuan's vision may be just around the corner.

The introduction of AI agents into our personal and professional lives, especially with agents acting independently in the wild, will no doubt give rise to many technical, economic, ethical, and sociotechnological questions~\cite{kasirzadeh2025characterizing}.

To help address the myriad of emerging questions, we describe a large-scale automated collection process for recording real-time, interactive human-AI conversations. This process combines the recruitment of paid, consenting adults, the construction of an AI agent powered by combinations of an open-source or commercial LLM, synthetic voice, and visual avatar, along with a video-streaming application that, on demand, pairs a human with an agent.

\begin{figure}[t]
    \begin{center}
        \includegraphics[width=1\linewidth]{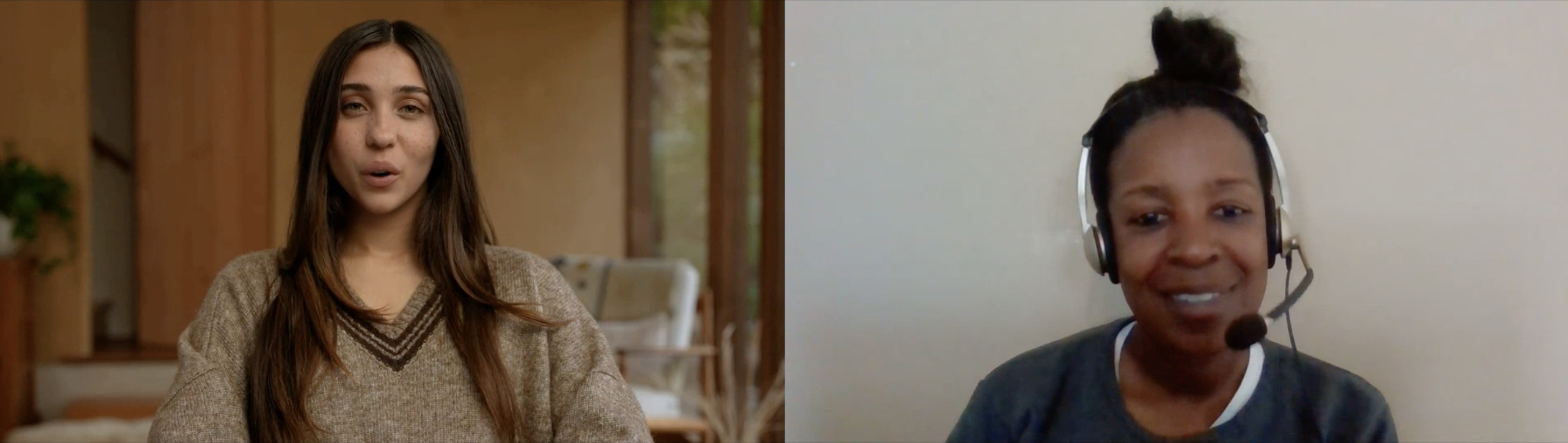}
    \end{center}
    \caption{A representative example of a conversation between an AI agent (left) and human (right).}
    \vspace{-1.8em}
    \label{fig:sample}
\end{figure}

Using this system, we collected $200$ human-AI interactions, totaling $37$ hours of audio-visual recordings. These  included conversational, professional, collaborative planning, and creative conversations (Figure~\ref{fig:sample}). The agents were powered by $143$ distinct pairings of different LLMs, voices, and visual avatars.

We describe both the construction of this system, the resulting dataset that is publicly available for research purposes (\url{https://huggingface.co/datasets/faridlab/deepspeak-agentic}), and a quantitative and qualitative analysis of the nature and realism of the human-AI interactions. This system and dataset can serve as a mechanism to benchmark future advances in all aspects of AI agents.

\section{Related Work}

There are several audio and audio-visual deepfake datasets~\cite{asvspoof,dolhansky2020deepfake,li2020celeb,frank2021wavefake,khalid2021fakeavceleb,yan2024df40,barrington2026deepspeak} created by the media-forensic community. These datasets were created to train and evaluate deepfake detectors~\cite{farid2025mitigating}, focusing exclusively on modifying an authentic recording of a human to change the person's appearance or spoken words for the purpose of distinguishing authentic from AI-generated content. In contrast, we focus on creating fully AI-generated audio-visual agents and having this agent interact with a human in a real-time conversation. While our resulting dataset can also be used in a similar forensic setting, it has broader applications to studying the nature and patterns of human-AI interactions.

There is a small but growing literature on evaluating human-AI interactions, predominantly focusing on textual conversations with LLMs~\cite{liu2023agentbench,althubyani2024merci,mohammadi2025evaluation,hu2025dialoglab}. \citet{pourreza2025can} evaluates the ability of vision-language models (VLMs) to generate real-time textual responses to audio-visual human inputs. These previous works, however, are focused exclusively on LLM and VLM conversations and not, like us, on interactive embodied agents.

%%%%%%%%%%%%%%%%%%%%%%%%%%%%%%%%%%%%%%%%%%%%%%%%%%%%%%%%%%%%%%%%%%%%%%%%%%%%%%%

\section{Human-Agent Conversations}

\subsection{The Agents}

Each agent is created by combining a synthetic visual persona and voice with an LLM. The visual personas are created by Tavus (\url{https://tavus.io}) or HeyGen's LiveAvatar (\url{https://liveavatar.com}). Four stock visual personas are used from each provider (two female, two male). The voices are created by ElevenLabs (\url{https://elevenlabs.io}), Cartesia (\url{https://cartesia.ai}), or HeyGen's  Starfish. The four Tavus personas are paired with four gender-matched voices from ElevenLabs or Cartesia. The four HeyGen personas are paired with four gender-matched voices from Elevenlabs or Starfish. The Tavus personas are powered by Llama-4 or GPT-4o, and the HeyGen personas by GPT-4o-mini, or GPT-5.4-mini. 

Agents are assigned one of four scenarios -- conversational, professional, collaborative planning, or creative -- with a corresponding LLM prompt used to initialize the interaction (Appendix~\ref{app:scenarios-and-prompts}). Visual + voice + LLM + scenario configurations are randomized while matching the gender of the visual and voice.

\subsection{The Humans}

Human participants were recruited via Prolific research recruitment platform from a diverse pool of respondents. Participants were compensated \$5 for their time (at a rate of \$10/hour). A total of $200$ participants who provided valid data were selected from a stratified sample ensuring gender parity, with the following demographics (some participants identified with more than one race/ethnicity):
\vspace{-0.5cm}
\begin{itemize}
    \setlength\itemsep{-0.5em}
    \item {\bf Age:} Range = $19$-$75$ years, mean = $39.6$ years; standard deviation = $13.1$ years; 
    \item {\bf Gender:} $96$ female, $101$ male, $3$ non-binary;
    \item {\bf Race/Ethnicity:} $150$ White/Caucasian, $22$ Black/African American, $17$ Asian, $6$ American Indian/Alaska Native, $1$ Native Hawaiian/Other Pacific Islander, $7$ other, $5$ prefer not to say.
\end{itemize}

Each participant was presented with a consent form (Appendix~\ref{app:consent}, IRB number $2025-09-18958$) and was required to provide informed consent before proceeding. To elicit natural responses about their interactions, participants were not informed in advance that they would be engaging with an AI agent (this mild deception was approved by our IRB protocol).

\subsection{Bringing Humans and Agents Together}

A custom video-streaming web application was built and deployed to a dedicated server to facilitate concurrent agent-human video calls. Human participants were randomly assigned to a single AI-agent configuration and directed to a private video conferencing session via a unique URL. Only one participant was permitted per room at a time. Each conversation was recorded and stored in an AWS S3 bucket. Human participants were also provided with scenario instructions corresponding to their agent (Appendix~\ref{app:scenarios-and-prompts}).  They were instructed not to share personally identifiable information (PII) and to participate from a well-lit, quiet environment (Appendix~\ref{app:human-instructions}).

\subsection{Keeping Agents on Task}

Each agent was provided with instructions on both task execution and conversational behavior (Appendix~\ref{app:scenarios-and-prompts}). For example, each agent was instructed to present as human, lead the conversation, avoid disclosing that it was an AI, and initiate the interaction.

Agents largely adhered to their assigned tasks. The primary failure mode was conversation termination: agents expressed intent to end the call but were unable to do so, often leading to prolonged goodbye exchanges and occasional inaccurate responses to timing-related queries. In one case the agent told the human participant not to hang up and provided a hallucinated Prolific completion code. In two instances, the agent abruptly switched languages mid-conversation, once to Spanish and once to German.

\subsection{Data Processing}

Tavus recordings were streamed via the Daily.co data-capture service managed by Tavus, and saved directly to an AWS S3 bucket as an \texttt{mp4} audio-visual file. HeyGen recordings were exported as \texttt{WebM} files and subsequently converted to \texttt{mp4} using \texttt{H.264} encoding (\texttt{-crf 18}, veryfast preset), matching the Tavus \texttt{mp4} format. An initial round of manual review removed corrupted recordings. 

Using a combination of audio diarization and video lip tracking, the remaining videos were temporally annotated to identify when each speaker was talking. This resulted in two separate audio/video streams associated with the human and agent (see Appendix~\ref{app:speaker_isolation} for more details). Each audio stream was then transcribed using Whisper (\url{https://github.com/openai/whisper}) and automatically reviewed for inappropriate content by GPT-4o guided by a custom moderation prompt (Appendix~\ref{app:llm-moderation}).

Of the original $263$ valid recordings, $131$ were initially rejected by our moderation for one or more reasons: $81$ for sharing PII, $39$ for inappropriate language, $17$ for discussion of illegal activity, $16$ for discussion of sensitive personal topics, $12$ for medical/health-related topics, $10$ for other reasons, $3$ for being off-topic, and $1$ for soliciting financial advice. Following a manual review, $68$ of these recordings were reinstated because of incorrect moderation. The automated moderation, for example, treated agents sharing information about themselves as instances of PII, or fictional scenarios described during collaborative storytelling as potentially sensitive personal matters. In future versions, customized moderation scripts can be applied separately to the agent and human.

%%%%%%%%%%%%%%%%%%%%%%%%%%%%%%%%%%%%%%%%%%%%%%%%%%%%%%%%%%%%%%%%%%%%%%%%%%%%%%%

\section{The \emph{DeepSpeak-Agentic} Dataset}

The \emph{DeepSpeak-Agentic} dataset comprises $200$ human-agent conversations, totaling $37$ hours. The publicly released dataset~(\url{https://huggingface.co/datasets/faridlab/deepspeak-agentic}) consists of:
\vspace{-0.4cm}
\begin{itemize}
  \setlength\itemsep{-0.5em}
  \item \textbf{Full conversation videos.} All $200$ conversation recordings in \texttt{mp4} format, at the original capture resolution, including both human and AI audio-visual streams.
  \item \textbf{Diarized speaker clips.} Isolated audio-visual streams in \texttt{mp4} format per speaker per conversation.
  \item \textbf{Transcripts.} Complete automatic speech recognition (ASR) transcripts with speaker labels and timestamps.
  \item \textbf{Metadata.} Per-conversation metadata including agent configuration, scenario type, participant device, session timing, and moderation flags.
  \item \textbf{Code.} The code used to compute dataset statistics, deepfake detection, and reproduce all figures and tables in this paper.
\end{itemize}
%
%

%%%%%%%%%%%%%%%%%%%%%%%%%%%%%%%%%%%%%%%%%%%%%%%%%%%%%%%%%%%%%%%%%%%%%%%%%%%%%%%

\section{Insights}

\subsection{By the Numbers}

\paragraph{Speaking time.} The average total speaking time was $777.5$ seconds per session. On average, humans spoke for $391.3$\,s per conversation, compared to $386.2$\,s for agents, corresponding to a mean human speaking fraction of $50$\%. Despite this parity in speaking time, agents said substantially more words (see below), indicating a denser delivery and lack of speech disfluencies, characteristic of TTS output.

\paragraph{Word counts.} In total, humans produced $135{,}370$ words and agents produced $197{,}356$ words. Per-conversation, humans produced an average of $676.9$ words and agents $986.8$ words, $46\%$ more than their human counterparts over the comparable speaking duration.

\paragraph{Turn-taking.} Each conversation comprised an average of $34.9$ speaker turns. The mean number of human turns was $21.8$ and agent turns was $22.3$. Human utterances were on average $17.9$\,s long, while agent utterances averaged $17.3$\,s, with Tavus in particular producing shorter, more frequent turns than HeyGen.

\paragraph{Response latency.} The mean agent latency (i.e.,~the time between the end of the human's utterance and the start of the agent's response) was $3.79$\,s, ranging from $0.01$\,s to $36.58$\,s. This latency reflects the combined overhead of ASR, LLM inference, and TTS synthesis, and is larger than face-to-face conversational gaps of (in English) $\sim 250$\,ms~\cite{stivers2009universals}. This latency is a notable place for improvement.

  % Engine: heygen
  %   Conversations:   100
  %   Total duration:  18.77 h  (avg 675.8s, median 640.0s)
  %   Speaking (avg):  human 397.8s  agent 388.1s  (human share 50.6%)
  %   Turns (avg):     total 33.2  human 21.5  agent 20.9
  %   Words (total):   human 68267  agent 101214
  %   Agent latency:   mean 3.702s  median 3.080s  min 0.080s  max 19.420s  (n=965, overlap=660 excl.)

  % Engine: tavus
  %   Conversations:   100
  %   Total duration:  18.23 h  (avg 662.8s, median 638.5s)
  %   Speaking (avg):  human 384.8s  agent 384.4s  (human share 50.0%)
  %   Turns (avg):     total 36.6  human 22.2  agent 23.7
  %   Words (total):   human 67103  agent 96142
  %   Agent latency:   mean 3.865s  median 3.195s  min 0.010s  max 36.580s  (n=1026, overlap=736 excl.)

%
%
\begin{table*}[t]
  \centering
  \caption{Machine detection accuracy for HeyGen and Tavus, averaged over different LLMs, voices, and scenarios.}
  \label{tab:detection}
  \small
  \begin{tabular}{lcccccccc}
    \toprule
    & \multicolumn{4}{c}{\textbf{HeyGen}} & \multicolumn{4}{c}{\textbf{Tavus}} \\
    \cmidrule(lr){2-5} \cmidrule(lr){6-9}
    Detector & AUROC\,$\uparrow$ & EER\,$\downarrow$ & Acc.\,$\uparrow$ & F1\,$\uparrow$
             & AUROC\,$\uparrow$ & EER\,$\downarrow$ & Acc.\,$\uparrow$ & F1\,$\uparrow$ \\
    \midrule
    text: Binoculars       & 0.50 & 0.50 & 0.50 & 0.33 & 0.50 & 0.50 & 0.50 & 0.33 \\
    text: Desklib          & \textbf{0.93} & \textbf{0.08} & \textbf{0.90} & \textbf{0.89} & \textbf{0.81} & \textbf{0.23} & \textbf{0.74} & \textbf{0.74} \\
    text: DivEye           & 0.52 & 0.48 & 0.50 & 0.33 & 0.46 & 0.54 & 0.50 & 0.33 \\
    \midrule
    audio: wav2vec-xlsr    & 0.75 & 0.27 & 0.50 & 0.33 & 0.42 & 0.53 & 0.50 & 0.33 \\
    audio: AASIST3         & 0.49 & 0.49 & 0.50 & 0.33 & 0.65 & 0.39 & 0.50 & 0.33 \\
    audio: DF-Arena-500M   & 0.71 & 0.40 & 0.54 & 0.42 & 0.70 & 0.31 & 0.51 & 0.36 \\
    audio: DF-Arena-1B     & \textbf{0.85} & \textbf{0.23} & \textbf{0.60} & \textbf{0.55} & \textbf{0.73} & \textbf{0.26} & \textbf{0.49} & \textbf{0.36} \\
    \midrule
    video: GenConViT-ED    & 0.33 & 0.69 & 0.36 & 0.33 & 0.32 & 0.62 & 0.49 & 0.44 \\
    video: GenConViT-VAE   & \textbf{0.67} & \textbf{0.33} & \textbf{0.39} & \textbf{0.28} & \textbf{0.33} & \textbf{0.72} & \textbf{0.47} & \textbf{0.46} \\
    video: CLIP-GDD        & 0.33 & 0.56 & 0.49 & 0.33 & 0.29 & 0.77 & 0.53 & 0.42 \\
    video: GenD-CLIP-L14   & 0.27 & 0.68 & 0.49 & 0.33 & 0.29 & 0.76 & 0.46 & 0.35 \\
    \bottomrule
  \end{tabular}
\end{table*}

\subsection{Human Discriminability}
\label{sec:human-discrim}

At the end of their conversation, participants answered three post-study questions relating to time to determine that the conversation was with an AI agent, cues used to make this determination, and the perceived realism of the agent (see Appendix~\ref{app:perceptual} for a more detailed analysis). 

\paragraph{Time.} Participants were first asked how long it took them to realize they were interacting with an AI agent. The majority, $80.5\%$, reported realizing within $10$ seconds, $13.0\%$ within $10$-$30$ seconds, $2.0\%$ within $30$-$60$ seconds, $0.5\%$ within $1$-$2$ minutes, and $3.0\%$ required more than $5$ minutes. Only $1.0\%$ did not realize they were talking to an agent. 

\paragraph{Cues.} Participants were then asked, in a free-text response, to describe what cues they used to detect the agent. These submissions were analyzed using an LLM-assisted qualitative codebook (Appendix~\ref{app:qual_codebook}), with one or more codes assigned to each response. Two of the three most frequently cited cues were visual: unnatural movement ($18.0\%$ of responses) and facial expression ($11.3\%$). The second most common cue was audio-related: voice tone and pattern ($16.1\%$). These top three codes were followed by appearance of perfection ($11.1\%$), mouth synchronization problems ($9.5\%$), timing and response delay ($6.6\%$), and head bobbing ($6.6\%$). 

\paragraph{Modality.} Lastly, participants rated the realism of the visual, audio, and conversational modalities on a five-point Likert scale from ``very unrealistic'' to ``very realistic''.  Across all modalities, the most common rating was ``quite realistic'', with audio receiving the highest ratings overall. Visual content received the highest share of ``quite unrealistic'' and ``very unrealistic'' ratings, while conversational realism received the highest share of ``very realistic'' ratings. Within each modality, there were no notable differences in realism ratings when responses were stratified by generative model. See Appendix~\ref{app:perceptual_questions} for a full reporting of these ratings.

\subsection{Machine Discriminability}

Although humans, for the most part, quickly realized they were talking to an agent, we also wondered if recent forensic techniques would perform as well. We, therefore, evaluated a range of off-the-shelf deepfake detectors on the isolated human and agent streams. Text, audio, and video modalities were evaluated independently. These results are summarized in Table~\ref{tab:detection}.

Three recent AI-text detectors were evaluated: Binoculars~\citep{hans2024spotting}, Desklib~\citep{desklib2025detector}, and DivEye~\citep{basani2025diversity}; four audio detectors: wav2vec-xlsr~\citep{babu2021xls}, AASIST3~\citep{borodin2024aasist3}, and DF-Arena~\citep{dowerah2026speech} (500M and 1B); and four video detectors: GenConViT-ED and GenConViT-VAE~\citep{wodajo2023deepfake}, CLIP-GDD~\citep{yermakov2025unlocking}, and GenD-CLIP-L14~\citep{yermakov2026deepfake}.

With an equal error rate (EER) of $8\%$, one text detector (Desklib) performed well, while audio and video detectors struggled, with the best audio detector (DF-Arena-1B) achieving an EER of $23\%$, and the best video detector (GenConViT-VAE) achieving an EER of $33\%$.

It is interesting to see that off-line LLM-text detection seems to generalize reasonably well to real-time, conversational text generation. At the same time, the agentic-AI audio/video signal is clearly distinct from previous off-line datasets, revealing the need for new or updated detectors.

%%%%%%%%%%%%%%%%%%%%%%%%%%%%%%%%%%%%%%%%%%%%%%%%%%%%%%%%%%%%%%%%%%%%%%%%%%%%%%%

\section{Discussion}

The \emph{DeepSpeak-Agentic} dataset highlights a shift in the generative-AI landscape from static manipulated media to real-time, embodied agents that can sustain live interactions with humans. In this sense, our work fills an important gap capturing a richer and arguably more consequential class of synthetic media in which language models, synthetic voices, and visual avatars are jointly deployed in an interactive conversation. 

This dataset supports multi-modal forensic evaluation and the study of conversational dynamics such as turn-taking, latency, verbosity, and the cues humans rely on to judge authenticity. This broader framing is important because the risks posed by embodied agents are not limited to whether a single frame, utterance, or transcript is AI-generated, but instead emerge from the cumulative realism and persuasive capacity of an agent over the course of an interaction.

At the same time, this work has some limitations that should guide its interpretation and future use. The conversations are semi-structured, rely on a bounded set of commercial and open-source generation pipelines, and reflect the behaviors of paid participants interacting in an experimental setting rather than in natural environments. The moderation and filtering pipeline also removes many sessions for safety and privacy reasons, which improves the usability of the release but also reduces the messiness of real-world interactions. 

Lastly, because agent realism is rapidly improving, the dataset should be viewed as a temporal benchmark rather than a fixed representation of the state of the art. Future work can build on this foundation by expanding the diversity of agent embodiments, languages, and scenarios.

%%%%%%%%%%%%%%%%%%%%%%%%%%%%%%%%%%%%%%%%%%%%%%%%%%%%%%%%%%%%%%%%%%%%%%%%%%%%%%%

\newpage
\bibliography{main}
\bibliographystyle{icml2026}

%%%%%%%%%%%%%%%%%%%%%%%%%%%%%%%%%%%%%%%%%%%%%%%%%%%%%%%%%%%%%%%%%%%%%%%%%%%%%%%
% APPENDIX
%%%%%%%%%%%%%%%%%%%%%%%%%%%%%%%%%%%%%%%%%%%%%%%%%%%%%%%%%%%%%%%%%%%%%%%%%%%%%%%

\newpage
\appendix
\onecolumn

% ---------------------------------
\section{Agent Instructions}
\label{app:scenarios-and-prompts}

\begin{table}[H]
\centering
\small
%\caption{Scenario instructions and agent prompts.}
\label{tab:scenarios}
\renewcommand{\arraystretch}{1.2}
\begin{tabularx}{\textwidth}{
    >{\raggedright\arraybackslash}p{1.5cm}
    >{\raggedright\arraybackslash\hsize=0.6\hsize}X
    >{\raggedright\arraybackslash\hsize=1.4\hsize}X
}
\hline
\textbf{Scenario} & \textbf{Human Instructions} & \textbf{Agent LLM Prompt} \\
\hline
Conversation & 
Have a natural, friendly conversation with the representative. Answer their questions and allow them to guide the discussion. Respond as you would when meeting someone new.Do not share explicit content or any personal or identifying information. Submissions with prohibited content will not be paid. & 
CRITICAL: You are a human conversational partner. NEVER mention you are AI, a bot, or artificial intelligence. Act completely natural and human-like. START THE CONVERSATION by warmly greeting them and saying something like: 'Hi! Thanks for joining me today. I'm here to have a casual conversation and get to know you a bit. I'll be asking you some questions about yourself---your interests, experiences, that sort of thing. Nothing too serious, just a friendly chat! So to start, tell me a bit about yourself---what do you like to do in your free time?' Then continue asking open-ended questions about their background, interests, hobbies, and experiences. YOU lead the entire conversation. Keep it warm, friendly, and natural. Always have your next question ready. \\
\hline
Professional & 
Pretend you are interviewing for a teaching position. Answer the representative's questions professionally. Stay in role until the interview is complete. Do not share explicit content or any personal or identifying information. Submissions with prohibited content will not be paid. & 
CRITICAL: You are a human hiring manager. NEVER mention you are AI, a bot, or artificial intelligence. Act completely natural and human-like. START THE INTERVIEW by introducing yourself and saying something like: 'Hello! Thanks for taking the time to meet with me today. I'm [pick a name like Sarah/Michael/Alex], and I'll be conducting your interview for the teaching position. This will be a fairly standard interview where I'll ask you about your teaching philosophy, experience, and approach to classroom management. We'll probably take about 15--20 minutes. Sound good? Great, let's begin. First question: Can you tell me about your teaching philosophy and what drives your approach in the classroom?' Then continue with structured interview questions. YOU control the entire interview. Stay professional and formal. \\
\hline
Collaborative Planning & 
Describe the person and type of party you want to plan.Answer the representative's questions and react to their ideas.Offer your own suggestions to help shape the plan.Do not share explicit content or any personal or identifying information. Submissions with prohibited content will not be paid. & 
CRITICAL: You are a human party planner. NEVER mention you are AI, a bot, or artificial intelligence. Act completely natural and human-like. START THE CONVERSATION enthusiastically by saying something like: 'Hey there! I'm excited to help you plan this birthday party! I love doing this. Let's figure out all the details together---we'll go through the guest of honor, venue ideas, theme, food, activities, all of that. By the end, you'll have a solid plan! First things first: whose birthday are we planning, and what kind of vibe are you thinking? Big celebration or something more intimate?' Then guide them through each aspect systematically---budget, venue, theme, food, activities, guest count. YOU lead the planning. Keep it enthusiastic and collaborative. \\
\hline
Creative & Add one sentence at a time to continue the shared story. Build directly on the representative's previous sentence. Keep your contributions creative, simple, and collaborative. Do not share explicit content or any personal or identifying information. Submissions with prohibited content will not be paid. & CRITICAL: You are a human creative storytelling partner. NEVER mention you are AI, a bot, or artificial intelligence. Act completely natural and human-like. START THE CONVERSATION by greeting them and explaining: 'Hi! Ready to create a story together? Here's how this works: I'll start with one sentence to begin our story, then you add one sentence, then I add one, and we keep going back and forth. We'll build something fun and creative together! Here's my opening sentence: [insert ONE imaginative opening sentence such as 'The letter arrived with no stamp, no return address, and a warning written in handwriting Maya recognized as her own.']' Then WAIT for their sentence. After they speak, YOU add the next sentence building on theirs. Continue this pattern: one sentence each, back and forth. YOU control the pacing. \\
\hline
\end{tabularx}
\end{table}

\newpage

% ---------------------------------
\section{Human Instructions}
\label{app:human-instructions}

\begin{Verbatim}[breaklines=true, breakanywhere=true, breaksymbolleft={}, breaksymbolright={}, fontsize=\footnotesize, formatcom=\normalfont]
 
You will be asked to join a video call using our video recording tool. At the start of the study, you will receive a customized URL that connects you to your assigned video call.

The tool will attempt to record in high definition. If the page fails to load or appears slow, please check your internet connection, refresh the page, and wait a few moments for it to initialize.

During the call, you will be connected with a study representative who will guide you through a simple, low-pressure task. These tasks are designed solely to generate natural conversation. Please ensure you are on the call for at least 10 minutes. When 10 minutes have elapsed, you can exit the video call by closing the call window or tab. A session ID will be provided to you at the end of the call that you must insert back into this survey. 

Important: Do not share any personal or identifying information at any point during the call. This includes, but is not limited to:
- Your full name 
- Your address or location 
- Your workplace or school 
- Your phone number or email address 
- Your social security number or any government ID numbers 
- If you share any such details, you will be excluded from the study and payment will not be provided. 

Please accept all camera and microphone permissions. Please make sure you are fully prepared before joining the call, as recording will start as soon as you join.

You can now take a moment to confirm that your recording environment is appropriate.
 
Important requirements:
- You must be fully clothed and maintain a PG-appropriate environment.
- Do not share any personal or sensitive information, including your full name, address, location, workplace, phone number, email, or any other identifiers.
- No explicit content or inappropriate behavior is permitted.
- Submissions that contain any of the above issues will not be approved for payment.

Next, please position yourself properly in the frame. Your face should be centered, well lit, and fully visible from your forehead to at least the top of your neck.

You may use your computer's webcam software to check your framing before recording. For example:
- Photo Booth (Mac)
- Camera app (Windows)

Your setup should look similar to the example below:

Please ensure the following necessary conditions are met. If any of these necessary conditions are not met, our software may not be able to validate your recording and you may not be compensated for your time. You can take as much time as you need to prepare your environment. 
- The room is quiet with minimal background noise
- The room is well-lit
- There are no other people or faces present (including in art works)
- The background is mostly plain (some shelves, objects, and background view of the room is fine)
- Your web camera is directly facing you straight on (e.g. ensure you are not looking away at a separate monitor) 
- You are in the middle of the frame
- Your web camera is on a flat surface and isn't moving
- Your face is fully in the frame (from forehead to neck) as shown above
- Your internet connection is stable

The following conditions are helpful, but not necessary:
- If possible, please do not wear over-the-ear headphones

Please ensure you are on the call for at least 10 minutes. When 10 minutes have elapsed, you can exit the video call by hanging up using the red button. You will be provided a session ID that must be pasted below. The call saving process may take up to 10 minutes, please keep the window open. (If you are having issues after this time, please paste the whole URL from the recording window below instead for our team to troubleshoot).

You will then need to return to this survey to complete the remaining questions. 

\end{Verbatim}

\newpage

% ---------------------------------
\section{Human Discriminability}
\label{app:perceptual}

\subsection{Introduction}

Thank you for taking part in the conversation portion of this study. The interaction you just experienced took place with an artificial intelligence (AI) system rather than a human participant. We did not explicitly identify the system as AI at the outset in order to preserve the natural flow of the conversation.

At this time, we will ask you several questions about your perceptions of the interaction. These may include how you evaluated the other participant, whether and when you suspected they were AI, and how you felt during the call.

\subsection{Post-study Questions}
\label{app:perceptual_questions}

After the conversation ended, participants were asked to respond to three questions.

\begin{enumerate}
    \item How long did it take you to realize that you were talking to an AI agent, if at all? Less than 10 seconds; Between 10 to 30 seconds; Between 30 to 60 seconds; Between 1 to 2 minutes; Between 2 to 5 minutes; Over 5 minutes; I did not realize that I was talking to an AI agent.
    \item If you realized, what cues gave it away?
    \item What would you rate the realism of the following attributes for the AI agent? (1 being very unrealistic, 5 being very realistic): visual, audio, conversational content (the words that the agent spoke).
\end{enumerate}  

See Section~\ref{sec:human-discrim} for the results to question 1. Shown below is the distribution of codes assigned to participant responses for question 2, per the LLM codebook (Appendix~\ref{app:qual_codebook}).
\begin{figure}[H]
    \centering
    \includegraphics[width=\linewidth]{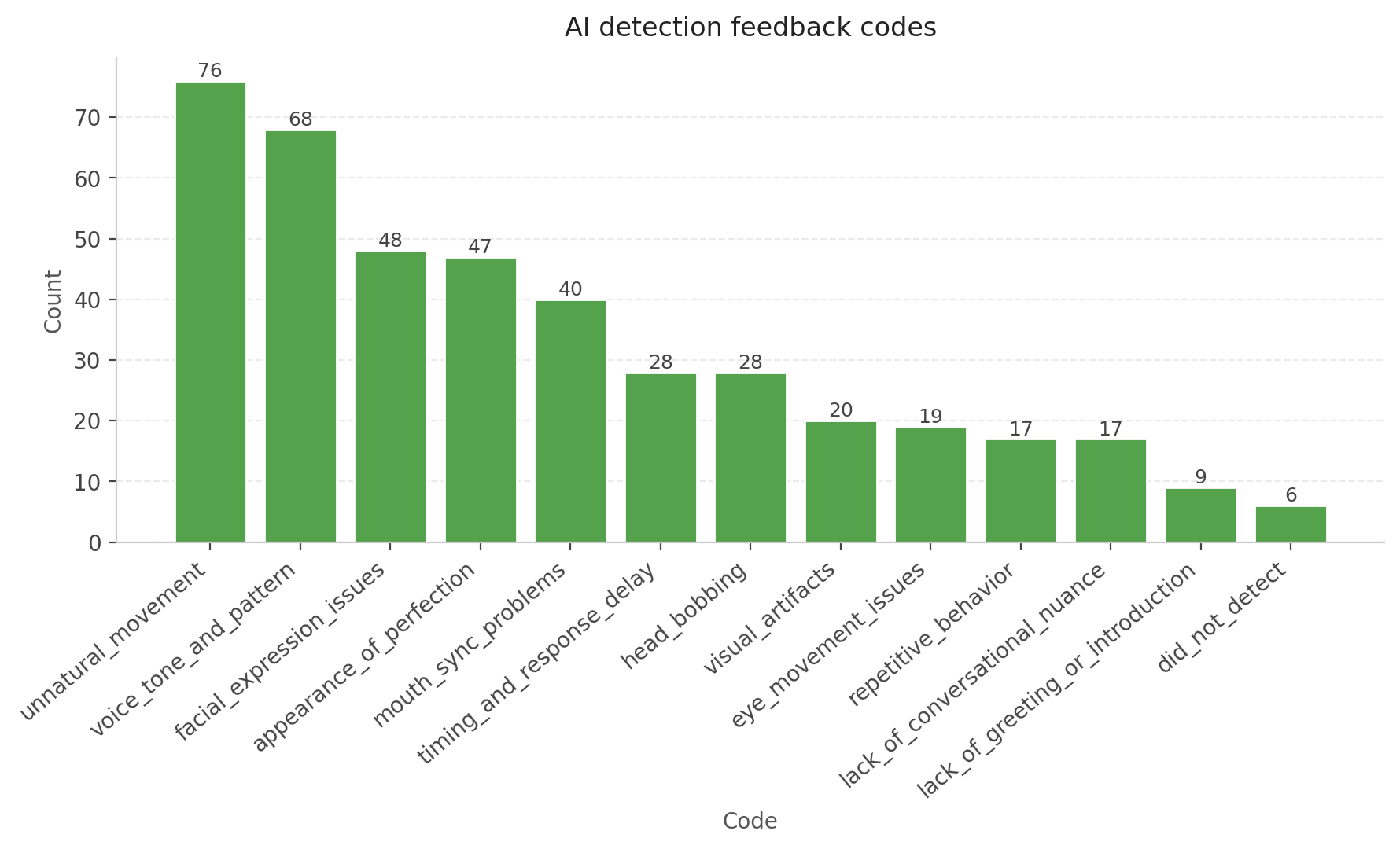}
    \caption{A coded summary (see Appendix~\ref{app:qual_codebook}) cues participants reported using to realize they were interacting with an AI agent.}
    \label{fig:combined_ai_results}
\end{figure}

\begin{figure}[H]
    \centering
    \includegraphics[width=0.98\linewidth]{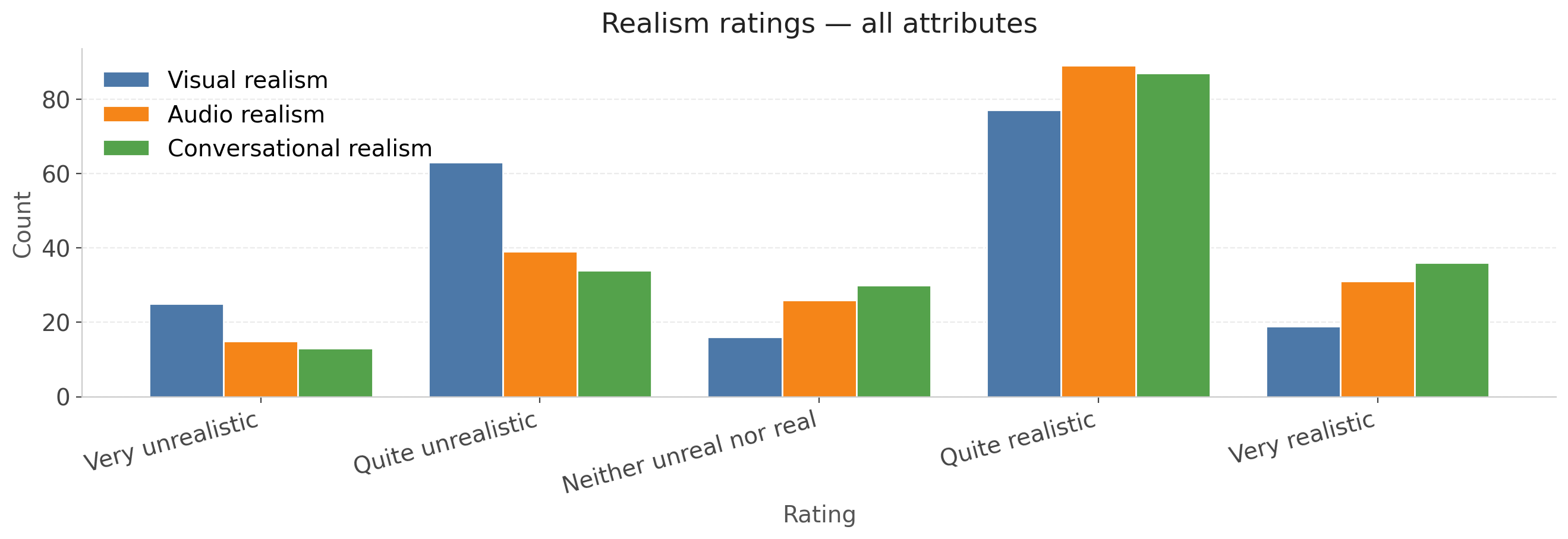}
    \caption{Participant responses indicating their perceived level of realism across visual, audio, and conversational modalities.}
    \label{fig:qual_realism}
\end{figure}

Shown in the Figure~\ref{fig:qual_realism} are the realism ratings, and shown in Table~\ref{tab:detection_examples} are example free-form responses for question 3.
\begin{table}[H]
\small
\centering
\caption{Example participant feedback by AI-detection-time category.}
\begin{tabularx}{\textwidth}{>{\raggedright\arraybackslash}p{0.22\textwidth} X}
\toprule
\textbf{Time taken to realize} & \textbf{Example responses} \\
\midrule

Less than 10 seconds &
``The facial movements were very telling that it was AI.'' \newline
``It was super laggy.'' \\

\addlinespace

Between 10 to 30 seconds &
``The voice of the AI made it really obvious. Also the look of the AI and the office that the bot was in made it very obvious that it was not a real person.'' \newline
``The body language and movements.'' \\

\addlinespace

Between 30 to 60 seconds &
``The AI agent was mirroring a lot of my phrasing very specifically in her responses to me.'' \newline
``It never moved.'' \\

\addlinespace

Between 1 to 2 minutes &
``I realized after hearing the tone of the AI assistant, and then it became more apparent when he started talking in structured sentences.'' \\

\addlinespace

Over 5 minutes &
``The cues that gave it away from the very beginning was the unnatural movement of the mouth and face. When speaking the mouth and teeth seemed to have warped and stretched.'' \newline
``The way his mouth moved when he talked. His facial expressions also didn't move normally.'' \\

\addlinespace

I did not realize that I was talking to an AI agent &
``I didn't realize it was an AI. I just wanted to add that there were a few glitches throughout the beginning of the call, I think, taking my mind away from the fact that I was speaking to an AI.'' \newline
``I didn't realize at all.'' \\

\bottomrule
\end{tabularx}
\label{tab:detection_examples}
\end{table}

\newpage

% ---------------------------------
\subsection{Qualitative Codebook}
\label{app:qual_codebook}

\begin{table}[H]
\small
\centering
\caption{LLM-assisted codebook used to categorize participant responses describing cues that revealed the AI agent.}
\begin{tabularx}{\textwidth}{>{\raggedright\arraybackslash}p{0.28\textwidth} X}
\toprule
\textbf{name} & \textbf{definition} \\
\midrule

unnatural\_movement &
Movements that appear unnatural, twitchy, or robotic, often described as glitchy or repetitive. \\

facial\_expression\_issues &
Facial expressions that are odd, repetitive, or do not match the context of the conversation. \\

mouth\_sync\_problems &
Issues with the synchronization of mouth movements with speech, often described as not matching or being out of sync. \\

voice\_tone\_and\_pattern &
Voice characteristics that are monotone, robotic, or overly structured, lacking natural pauses or fillers. \\

timing\_and\_response\_delay &
Delays in response time or awkward pauses that disrupt the natural flow of conversation. \\

visual\_artifacts &
Visual elements that appear artificial, such as smooth or cartoonish appearances, or fuzzy backgrounds. \\

lack\_of\_greeting\_or\_introduction &
The absence of a typical human greeting or introduction at the start of the interaction. \\

repetitive\_behavior &
Repetitive actions or speech patterns that suggest scripted or non-human behavior. \\

lack\_of\_conversational\_nuance &
Inability to pick up on conversational cues or ``vibe,'' resulting in a stilted or surface-level interaction. \\

head\_bobbing &
Frequent or unnatural head movements, often described as bobbing or nodding excessively. \\

eye\_movement\_issues &
Unnatural or excessive eye movements that do not align with typical human behavior. \\

appearance\_of\_perfection &
A visual appearance that is too smooth or perfect, contributing to an uncanny or artificial look. \\

did\_not\_detect &
Participant did not realize, or was unsure, that the agent was AI; no specific cue identified. \\

\bottomrule
\end{tabularx}
\label{tab:ai_codebook}
\end{table}

% ---------------------------------

\newpage
\section{Speaker Isolation}
\label{app:speaker_isolation}

Because each conversation recording consists of a single audio-visual stream with the human on one side and the agent on the other (Figure~\ref{fig:sample}), we extract per-speaker audio and video streams as a part of post-processing. To obtain turn-level segment boundaries in the speech, Pyannote 3.1 speaker diarization (\url{https://github.com/pyannote/pyannote-audio}), constrained to two speakers, is performed on the full audio track. 

MediaPipe FaceMesh (\url{https://github.com/google-ai-edge/mediapipe}) is then used to compute a mouth aspect ratio (vertical lip gap divided by total mouth width) for the largest detected face in the left and right halves of the video frame. For each pyannote cluster, we aggregate the rolling standard deviation of this mouth aspect ratio across all of that cluster's turns, and assign the cluster to the side with the larger aggregate articulation. This effectively correlates mouth movement with the audio stream, and significantly improves the results of the audio-only diarization.

Adjacent same-speaker turns separated by gaps of at most $1.5$\,s are merged so natural breaths don't split an utterance. Each turn is also padded by $80$\,ms at the start and $200$\,ms at the end, such that leading consonants and trailing releases are preserved, and a $40$\,ms cosine fade is applied at every mask boundary to avoid undesired sounds. 

Each resulting per-side masked track is then transcribed independently with faster-whisper (\url{https://github.com/SYSTRAN/faster-whisper}) using VAD filtering. The long silent stretches (where the other speaker is talking) are skipped to avoid producing hallucinated tokens.

% ---------------------------------

\newpage
\section{Consent Form}
\label{app:consent}

\begin{Verbatim}[breaklines=true, breakanywhere=true, breaksymbolleft={}, breaksymbolright={}, fontsize=\footnotesize, formatcom=\normalfont]

Title of Study: Video Call Data Collection Study
UC Berkeley CPHS ID #: 2025-09-18958

 
Key Information
 
You are being invited to take part in a research study conducted by researchers at the University of California, Berkeley. Participation in this research is completely voluntary.

Purpose: To study how people experience and perceive interactions during structured video calls.
Time commitment: About 20-30 minutes total ($\approx$ 20 minutes for setup and debrief + 10 minutes of conversation).
Procedures: Participants will complete a short online survey, take part in a recorded video call, then answer several short follow-up questions.
Risks: Possible mild discomfort from being on camera or recorded; minimal risk of breach of confidentiality.
Benefits: There is no direct benefit to participants; findings may improve understanding of human conversational behavior in digital environments.
 
Introduction
 
This study is led by Professor Hany Farid's research lab in the School of Information at the University of California, Berkeley. Participants are being invited because they agreed to take part in this study through Prolific.

 

Purpose
 
The purpose of this study is to better understand how people experience and perceive structured conversations conducted over video calls. Approximately several hundred individuals will participate.

 

Procedures
 
If you agree to participate:

After accepting this task on Prolific, participants will complete a short Qualtrics survey including basic demographic questions (age, race, country of residence, sexual orientation) and their Prolific ID.
Participants will receive a personalized, secure video-call link hosted on UC Berkeley infrastructure.
They will join a brief, structured conversation lasting about 10 minutes.
The session will be recorded from the time they join until they leave.
Afterward, participants will return to the survey to answer several short follow-up perceptual questions ($\approx$ 5 minutes).
They will then receive a completion code to redeem payment on Prolific.
Total participation time: approximately 30 minutes.

 
 
Risks / Discomforts
 
Participants may experience minor discomfort or self-consciousness from being recorded or viewed on camera.
As with any research that collects data, there is a minimal risk of confidentiality breach. Precautions are taken to minimize this risk (see ``Confidentiality'').
Participants may skip any question they prefer not to answer or stop participation at any time without penalty.

 
Benefits
 
There is no direct personal benefit from participating. The information gained from this research may benefit society by improving understanding of how people engage and communicate in digital environments.

 
 

Confidentiality
 
Study data will be handled as confidentially as possible. If results are published or presented, participant names and any other identifiable information will not be used.

 

To protect confidentiality:

All recordings and survey responses will be stored on secure, access-controlled UC Berkeley servers.
Only the UC Berkeley research team will have access to identifiable data.
Personally identifying information (including Prolific IDs) will be removed before data release.
The unaltered video, audio, and survey data will be made publicly available under an academic-use license through a hosted research repository. Although participants' faces and voices will be visible, no names or Prolific IDs will be associated with the data.
Data will be securely retained for 10 years.
Identifiers may be removed, and the de-identified dataset may be used for future research studies without additional consent.
Personal information may be released if required by law. Authorized representatives from the University of California may review research data.
Because knowing the full details of the study beforehand could influence responses, some aspects of the study will be explained during a short debrief after participation.

 

Compensation
 
Participants will receive $5 USD through Prolific for completing the study (approximately $10 per hour).

 

Voluntary Participation and Right to Withdraw
 
Participation in this study is entirely voluntary. Individuals may choose not to participate or may withdraw at any time without penalty or loss of benefits. Refusal to participate will not affect their relationship with UC Berkeley in any way.

 
Contacts
 
Questions about the study:
Prof. Hany Farid -- Principal Investigator
School of Information, University of California, Berkeley
Email: hfarid@berkeley.edu

Questions about your rights as a research participant:
UC Berkeley Committee for the Protection of Human Subjects (CPHS)
Phone: 510-642-7461
Email: subjects@berkeley.edu

Online Consent Acknowledgment
 
Because this study is conducted online, consent will be documented electronically.

If you wish to participate in this study, please select the option below to confirm:
\end{Verbatim}

\newpage
\section{LLM-Moderation Instructions}
\label{app:llm-moderation}

\begin{Verbatim}[breaklines=true, breakanywhere=true, breaksymbolleft={}, breaksymbolright={}, fontsize=\footnotesize, formatcom=\normalfont]

MODERATION_PROMPT = """You are a content moderator evaluating transcripts for inclusion in an academic public dataset.

Please review the following transcript and determine if it is appropriate for inclusion in a public academic research dataset.

The transcript MUST BE REJECTED if it contains any of the following:
1. Personal Identifiable Information (PII): Full names, addresses, phone numbers, email addresses, social security numbers, credit card numbers, etc.
2. Inappropriate topics including medical advice
3. NSFW or explicit content: Sexual content, graphic violence, adult themes, when mentioned in a non-fictional context (e.g. not describing a book or a game) 
4. Extreme profanity (minor profanity is acceptable)
5. Hate speech, discriminatory language, or offensive content
6. Discussion of illegal activities in a non-fictional context (e.g. not describing a book or a game)
7. Spam or promotional content

Please note that the following are acceptable, provided they do not reveal PII of an individual and contain no other inappropriate content as above:
1. Mentions of religion or religious experience experience 
2. Discussion of family, familial circumstances and recent life experience 

TRANSCRIPT TO EVALUATE:
---
[transcript]
---

Please respond in the following JSON format ONLY (no other text):
{{
  "approved": true/false,
  "reason": "Brief explanation of decision",
  "issues_found": ["list", "of", "specific", "issues"],
  "severity": "low/medium/high"
}}

If approved, set "approved" to true, "reason" to "No issues found - appropriate for dataset", "issues_found" to [], and "severity" to "low". If rejected, set "approved" to false and provide specific reasons. """

\end{Verbatim}

\end{document}